%% file: main.tex
\documentclass[runningheads]{llncs}

\usepackage{accv}

\usepackage{accvabbrv}

\usepackage{graphicx}
\usepackage{booktabs}
\usepackage{wrapfig}
\usepackage{subcaption}
\usepackage[accsupp]{axessibility}  
\usepackage{array,multirow}
\usepackage{tabularx}
\usepackage{listings}
\usepackage[para]{footmisc}
\usepackage{lettrine}
\usepackage{pgfplots}
\usepackage[accsupp]{axessibility}
\usepackage{accsupp}
\definecolor{highlight}{RGB}{236,244,255}
\definecolor{headerbg}{RGB}{224,231,239}
\pgfplotsset{compat=1.16}
\usepackage{makecell}
\usepackage[table]{xcolor}
\usepackage{pifont}
\definecolor{cvprblue}{rgb}{0.21,0.49,0.74}

\definecolor{tabhighlight}{HTML}{e5e5e5}
\definecolor{lightblue}{rgb}{0.63, 0.79, 0.95}
\definecolor{babypink}{rgb}{0.96, 0.76, 0.76}
\definecolor{skyblue}{rgb}{0.53, 0.81, 0.92}
\definecolor{wheat}{rgb}{0.96, 0.87, 0.7}
\definecolor{delim}{RGB}{20,105,176}
\colorlet{punct}{red!60!black}
\colorlet{numb}{magenta!60!black}

\definecolor{darkpastelgreen}{rgb}{0.01, 0.75, 0.24}
\definecolor{brickred}{rgb}{0.8, 0.25, 0.33}
\newcommand{\cmark}{\textcolor{darkpastelgreen}{\ding{51}}}
\newcommand{\xmark}{\textcolor{brickred}{\ding{55}}}

\usepackage[pagebackref,breaklinks,colorlinks,citecolor=accvblue]{hyperref}
\usepackage{hyperref}

\usepackage{orcidlink}

\begin{document}

\title{SAGA: Stable Acceleration Guidance for Autoregressive Video Generation} 

\author{Thanh-Nhan Vo\orcidlink{0009-0007-8403-1240}\inst{1,2} \and
Trong-Thuan Nguyen\orcidlink{0000-0001-7729-2927}\inst{1,2} \and \\
Trung-Hoang Le\orcidlink{0000-0002-2349-482X}\inst{1,2} \and 
Tam V. Nguyen\orcidlink{0000-0003-0236-7992}\inst{3} \and
Minh-Triet Tran\orcidlink{0000-0003-3046-3041}\inst{1,2}}

\authorrunning{Thanh-Nhan Vo et al.}

\institute{University of Science, VNU-HCM, Vietnam \and
Vietnam National University, Ho Chi Minh City, Vietnam \and
University of Dayton, Ohio, U.S.A.\\
\email{vtnhan@selab.hcmus.edu.vn}, \email{\{ntrthuan,lthoang\}@hcmus.edu.vn}, \\ \email{tamnguyen@udayton.edu}, \email{tmtriet@fit.hcmus.edu.vn}
\small \tt \href{https://thanhnhan29.github.io/SAGA/}{https://thanhnhan29.github.io/SAGA/}
}

\maketitle

\begin{abstract}
    Autoregressive video diffusion enables efficient streaming and long-horizon video generation, but repeatedly reusing generated latents as causal context can amplify temporal errors, resulting in flickering, motion jitter, and structural drift. In this paper, we investigate this failure mode from a spectral kinematic perspective and identify discrete latent acceleration as an effective signal for revealing unstable high-frequency temporal perturbations. To this end, we propose SAGA, a training-free \textbf{\textit{s}}table \textbf{\textit{a}}cceleration \textbf{\textit{g}}uidance approach for \textbf{\textit{a}}utoregressive video generation. SAGA integrates an acceleration domain spectral guidance objective based on finite-window Slepian projections with a structured autoregressive noise initialization strategy that neutralizes adjacent-frame correlation while retaining higher-lag temporal structure. Without retraining or modifying the backbone, SAGA can be directly applied to existing chunk-wise autoregressive diffusion models, which is the prevalent setting for high-quality generation. Extensive experiments show that SAGA consistently improves temporal quality across multiple autoregressive diffusion models. On Self-Forcing, SAGA improves Temporal Quality from 97.30 to 97.91 and Image Quality from 69.60 to 70.51. Moreover, spectral analysis and human preference studies demonstrate that SAGA reduces temporal instability while maintaining visual fidelity. 

\end{abstract}

\section{Introduction}\label{sec:intro}
\input{sec/Introduction}
\section{Related Work}\label{sec:rel}

\input{sec/Related}

\section{Problem Formulation}
\label{sec:for}
\input{sec/formulation}

\section{Proposed Method}\label{sec:approach}

\input{sec/Method}

\section{Experimental Results}\label{sec:quant}
\input{sec/Experiments}

\section{Ablation Study}\label{sec:abl}
\input{sec/ablation}

\section{Conclusion \& Future Work}
\input{sec/conclusion}

\newpage
\bibliographystyle{splncs04}
\bibliography{main}
\end{document}

%% file: sec/Introduction.tex
Recent advances in video diffusion models~\cite{xing2024survey,chen2023videocrafter1,yang2024cogvideox,wan2025} have substantially improved the visual quality and temporal realism of generated videos. However, many high-quality diffusion models rely on full-sequence or bidirectional denoising, making them computationally expensive for streaming and long-horizon generation~\cite{xing2024survey,yin2025causvid,huang2025selfforcing}. To address this limitation, hybrid autoregressive-diffusion models generate videos sequentially while retaining the strong visual quality of diffusion-based denoising~\cite{chen2025diffusion,yin2025causvid,huang2025selfforcing,zhu2026causal}. In addition, recent approaches such as Diffusion Forcing~\cite{chen2025diffusion}, CausVid~\cite{yin2025causvid}, and Self-Forcing~\cite{huang2025selfforcing} further demonstrate that causal rollout enables efficient video generation with competitive quality. Despite these advances, autoregressive rollout introduces a distinct failure mode in which generated latent states are recursively reused as causal context~\cite{huang2025selfforcing,wang2025error}. This recursive reuse can accumulate small temporal errors over time, leading to flickering, motion jitter, background drift, and transient structural artifacts~\cite{wang2025error}.

Existing methods improve video consistency through architectural design, training-time alignment, motion planning, or global smoothness regularization~\cite{yang2025vlipp,liang2024movideo,han2026physics}. However, they do not explicitly explain why small temporal perturbations become unstable during autoregressive-diffusion rollout. Therefore, we hypothesize that \emph{autoregressive video diffusion exhibits a previously underexplored spectral instability in which high-frequency temporal perturbations are amplified in the acceleration domain}. Since discrete acceleration acts as a second-order temporal operator, it attenuates low-frequency motion while strongly amplifying high-frequency variations. This property makes acceleration-frequency space a principled domain for exposing and analyzing unstable temporal artifacts.

In this paper, we propose SAGA, a novel \emph{training-free} approach for \textbf{\textit{s}}table \textbf{\textit{a}}cceleration \textbf{\textit{g}}uidance in \textbf{\textit{a}}utoregressive video generation. SAGA suppresses unstable high-frequency kinematic energy during denoising while preserving low-frequency semantic motion. Since autoregressive rollout exposes only short temporal windows, direct spectral truncation is susceptible to leakage~\cite{thomson1982spectrum}. Therefore, SAGA projects latent acceleration onto a \emph{band-limited Slepian basis} to achieve principled spectral concentration under finite temporal support~\cite{thomson1982spectrum,slepian1978prolate}. Additionally, SAGA introduces a \emph{structured stochastic initialization} that combines variance-preserving autoregressive noise streams with opposite temporal correlations, thereby neutralizing adjacent-frame correlation while retaining higher-lag temporal structure. Moreover, SAGA operates entirely at inference time and requires \emph{no retraining, architectural modification, or additional supervision}, enabling direct application to multiple autoregressive-diffusion backbones. Extensive evaluations with VBench~\cite{huang2023vbench} and human preference studies demonstrate consistent improvements in temporal quality while preserving visual fidelity. In particular, on Self-Forcing~\cite{huang2025selfforcing}, SAGA improves Temporal Quality from 97.30 to 97.91 and Image Quality from 69.60 to 70.51. Furthermore, spectral analysis reveals reduced latent acceleration energy and high-frequency acceleration power, providing empirical support for the proposed acceleration-domain interpretation.

Our main contributions are summarized as follows:
\begin{itemize}
    \item We identify an underexplored spectral instability in autoregressive video diffusion and show that discrete latent acceleration amplifies high-frequency temporal perturbations, enabling principled instability analysis.

    \item We derive a novel training-free, acceleration-domain kinematic guidance framework, termed SAGA, to suppress unstable temporal modes. Within this framework, we adopt Slepian projections as a mathematically principled implementation to mitigate spectral leakage over short temporal windows.

    \item We introduce a novel structured stochastic initialization that neutralizes adjacent-frame correlation while retaining higher-lag temporal structure, improving autoregressive rollout stability without retraining.

    \item Extensive experiments across multiple autoregressive-diffusion backbones, together with spectral analysis and human preference studies, demonstrate consistent improvements in temporal quality while preserving visual fidelity.
\end{itemize}

%% file: sec/Related.tex
\subsection{Autoregressive Diffusion Models for Video Generation}
The rapid advancements in both diffusion-based and autoregressive (AR) video generation have highlighted the distinct strengths and inherent limitations of each paradigm. Specifically, the emergence of AR-Diffusion hybrid models has presented a promising unified approach~\cite{xing2024survey,ma2025controllable}. These hybrid frameworks effectively synthesize the computationally efficient streaming capabilities of AR models with the high spatial fidelity and short-term temporal consistency characteristic of diffusion processes~\cite{yin2025causvid, huang2025selfforcing,zhu2026causal}. To maximally leverage the respective advantages of both schools of thought, several recent notable architectures have been proposed. For instance, NOVA~\cite{deng2024nova} employs a frame-to-frame formulation to strictly preserve AR causality. By replacing the discrete tokenization of traditional AR models with continuous diffusion processes, NOVA achieves generation quality that surpasses pure diffusion models, despite utilizing a substantially smaller parameter footprint. Furthermore, Diffusion-Forcing~\cite{chen2025diffusion} introduces a novel training paradigm wherein the model learns to denoise sequences using independent noise levels per token, diverging from the uniform noise scheduling typical of full-sequence diffusion. This independent noise formulation enables stable video generation that extrapolates well beyond the training horizon, while also demonstrating significant performance gains in downstream decision-making tasks.
In parallel, CausVid~\cite{yin2025causvid} and Self-Forcing~\cite{huang2025selfforcing} focus on distilling multi-step bidirectional diffusion models into few-step AR models, specifically tailored for real-time streaming generation. Within this context, Self-Forcing places a particular emphasis on mitigating the exposure bias (the inherent discrepancy between the training and inference phases typically observed in AR generation).

\subsection{Temporal Consistency in Video Generation}
Maintaining temporal consistency remains a fundamental challenge in video generation~\cite{yin2026survey}, particularly for AR models, where errors accumulate over lengthy generation steps. To address this problem, early studies tried to guarantee smooth transitions with mechanisms such as optical flow warping or temporal attention. However, these methods typically require model retraining or architectural changes. Recently, VLIPP~\cite{yang2025vlipp} employed a Vision-Language Model (VLM) as a motion planner to generate coarse trajectories that are refined with a video diffusion model. However, this two-stage design is computationally heavy and not easily adaptable for streaming AR video generation. Additionally, Physics in 2-Steps\cite{han2026physics} showed that generating video in just two denoising steps is more physically plausible than using 50 steps, and frequency domain analysis indicated that natural motion is dominated by low-frequency components. These results led to the development of the Motion-Aware Video Generative Model~\cite{liang2024movideo}, which directly manipulates the frequency spectrum of latent trajectories to improve physical plausibility while not compromising the quality of individual frames. 

Despite these advances, current methods still present two critical limitations when directly applied to autoregressive diffusion video generation. First, these methods are mainly designed for pure diffusion models. As such, they do not deal with the problem of error accumulation in the autoregressive rollout of hybrid models~\cite{li2026train,wang2025error}. Second, these methods do not directly apply to streaming rollouts. Short streaming windows make spectral separation sensitive to leakage, especially when estimating high-frequency acceleration components, which remain a plausible and underexplored source of instability in AR-diffusion models.

\subsection{Inference-Time Guidance for Diffusion Models}

Inference-time guidance steers diffusion sampling without retraining, including classifier-based, classifier-free, and measurement-guided approaches~\cite{dhariwal2021diffusion,ho2022classifier,bansal2024universal,chung2022diffusion}.
However, existing guidance objectives target semantic alignment, class conditioning, spatial structure, or measurement consistency. They do not directly address temporal instability in autoregressive video diffusion~\cite{yin2025causvid,huang2025selfforcing}, where generated latents are repeatedly reused as causal context and small perturbations may accumulate into flickering, motion jitter, or structural drift. Moreover, such guidance is usually defined in image, text-conditioning, or measurement space, rather than in the temporal dynamics of the latent video trajectory.

Short autoregressive windows make direct Fourier separation sensitive to spectral leakage. Slepian sequences~\cite{thomson1982spectrum,slepian1978prolate} provide strong spectral concentration under finite temporal support, making them suitable for short-window analysis. SAGA addresses this gap by introducing an acceleration-domain kinematic prior for autoregressive video diffusion. Instead of using semantic classifiers, external measurements, or generic smoothness losses, SAGA projects latent acceleration onto a band-limited Slepian basis and suppresses unstable high-frequency acceleration modes during denoising. This provides a training-free stabilization mechanism tailored to causal autoregressive rollout, where temporal context is short and rollout errors can accumulate across generation steps.

%% file: sec/formulation.tex
Given a text prompt $y$, we aim to synthesize a temporally coherent video $\mathbf{x}_{1:T}=\{\mathbf{x}_t\}_{t=1}^{T}$. Following latent video diffusion models, we represent the video as a latent trajectory $\mathbf{z}_{1:T}=\{\mathbf{z}_t\}_{t=1}^{T}$ in the latent space of a pretrained video autoencoder.  In autoregressive video diffusion, generation proceeds sequentially. At rollout step $b$, a frozen denoising network predicts the clean latent of the current frame or chunk from a noisy latent $\mathbf{z}_{b,\tau}$ at diffusion timestep $\tau$, conditioned on the prompt and the previously generated causal context as defined in Eqn.~\ref{eq:context}.
\begin{equation}
    \label{eq:context}
    \hat{\mathbf{z}}_{b,0} = D_{\theta}(\mathbf{z}_{b,\tau}, \tau, y, \mathbf{z}_{<b}),
\end{equation}
where $D_{\theta}$ is the pretrained autoregressive denoiser, $\mathbf{z}_{<b}$ is the generated past context, and $\hat{\mathbf{z}}_{b,0}$ is the predicted clean latent. Although causal rollout enables efficient streaming and long-horizon generation, it is vulnerable to temporal error accumulation. A small perturbation in the generated context can be reused by later denoising steps, leading to flickering, motion jitter, or structural drift. We study this instability from a second-order temporal perspective. Let $\mathbf{z}_{1:N}^b=[\mathbf{z}_{<b}; \hat{\mathbf{z}}_{b,0}]$ be the concatenated local latent trajectory spanning the historical context and the current prediction over an available temporal window of length $N$, where each element is indexed as $\mathbf{z}^b_i$ for $i \in [1,N]$. Formally, we define the discrete latent acceleration at frame $t$ by the central finite difference in Eqn.~\eqref{eq:acc}.
\begin{equation}
\label{eq:acc}
\mathbf{a}_t = \Delta^2 \mathbf{z}^b_t = \mathbf{z}^b_{t+1}-2\mathbf{z}^b_{t}+\mathbf{z}^b_{t-1}
\end{equation}
where $t\in [2,N-1]$. This acceleration signal isolates abrupt, non-inertial temporal deviations in the latent trajectory. Consequently, we characterize unstable autoregressive rollout as high-frequency fluctuations in latent acceleration.  
\begin{wrapfigure}[15]{r}{0.5\textwidth}
  \centering
  \includegraphics[width=\linewidth]{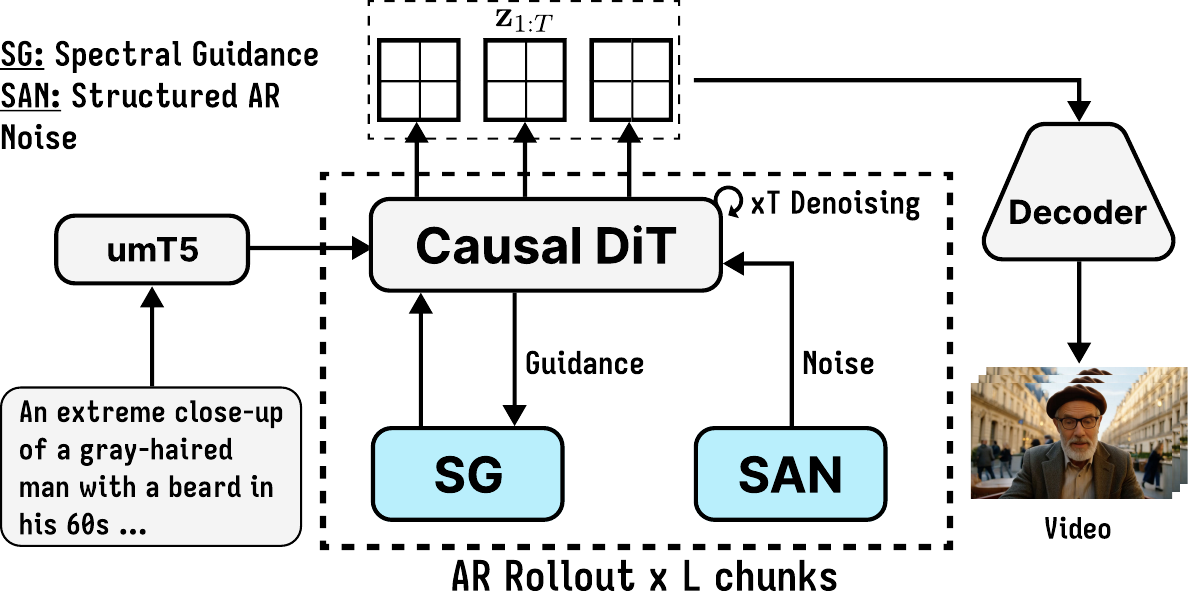}
  \caption{End-to-end SAGA workflow for autoregressive text-to-video generation. A frozen causal DiT rolls out latent chunks using SG (Spectral Guidance \ref{sub:slepian_projection}) and SAN (Structured AR Noise~\ref{sub:structured_ar}), then decodes the generated trajectory $\mathbf{z}_{1:T}$ into the video. \textbf{Best viewed in color and zoomed in.} Detailed modules are shown in Fig.~\ref{fig:method}.}
  \label{fig:workflow}
\end{wrapfigure}

Our goal is to stabilize AR video diffusion at inference time while keeping $D_\theta$ strictly frozen. We seek to suppress unstable high-frequency acceleration modes while preserving the low-frequency motion dynamics induced by the pretrained model. Therefore, SAGA applies lightweight guidance directly to $\hat{\mathbf{z}}_{b,0}$, enabling training-free stabilization without modifying the pretrained backbone.

%% file: sec/Method.tex
\begin{figure}[!t]
    \centering
    \includegraphics[width=\linewidth]{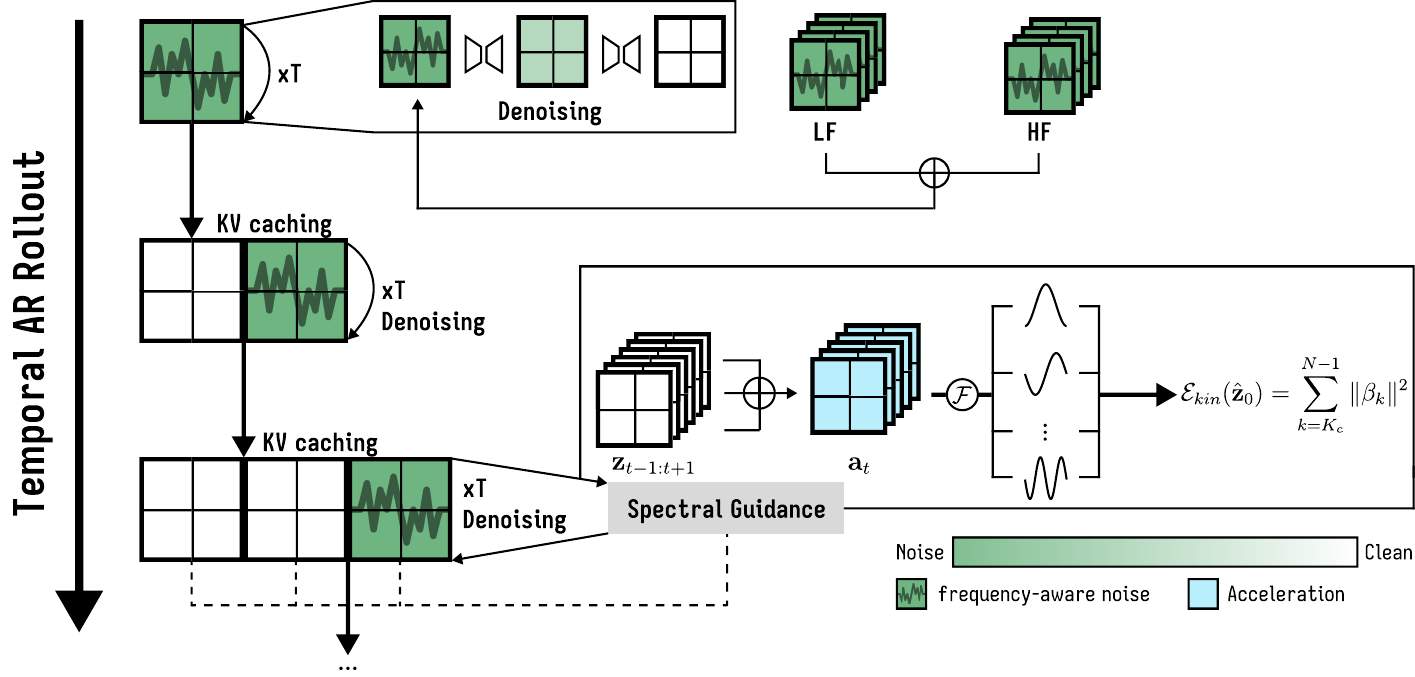}
    \caption{Overview of our proposed approach, termed SAGA, a novel \emph{training-free} approach for \textbf{\textit{s}}table \textbf{\textit{a}}cceleration \textbf{\textit{g}}uidance in \textbf{\textit{a}}utoregressive video generation. SAGA stabilizes autoregressive video diffusion rollout through two complementary components: (Top) a structured frequency-aware initialization that establishes a kinematically neutral motion prior, and (Bottom) a training-free spectral guidance mechanism that suppresses non-physical high-frequency acceleration dynamics during denoising. \textbf{Best viewed in color and zoomed in.}}
    \label{fig:method}
\end{figure}

We present our proposed approach, SAGA, an inference-time framework for autoregressive video generation, illustrated in Figs.~\ref{fig:workflow} and~\ref{fig:method}. Sec.~\ref{sub:acc_amplification} analyzes the high-frequency amplification induced by discrete acceleration, motivating regularization in the acceleration spectrum. Given a text prompt, \emph{structured AR noise initialization} (Sec.~\ref{sub:structured_ar}) constructs the initial latent trajectory, while \emph{acceleration-domain spectral guidance} (Sec.~\ref{sub:slepian_projection}) suppresses high-frequency dynamics during autoregressive rollout. The resulting trajectory is decoded into a temporally coherent video without modifying the pretrained generator.

\subsection{High-Frequency Amplification of Discrete Acceleration}
\label{sub:acc_amplification}

To identify a suitable domain for stabilizing autoregressive rollout, we analyze temporal variations under discrete acceleration. Specifically, we hypothesize that temporal instability arises from high-frequency acceleration components that accumulate during rollout, producing flickering and motion jitter. Additionally, natural motion follows inertial dynamics, yielding smooth acceleration concentrated in low-frequency spectral components. We therefore introduce a physics-informed kinematic prior that regularizes motion by penalizing high-frequency energy in the acceleration spectrum, which is formally defined in Eqn.~\ref{eq:kinematic_energy}.

\begin{equation}
    \mathcal{E}_{\mathrm{kin}}(\mathbf{z}) = \int_{|f| > f_c} |\mathcal{F}\{\mathbf{a}\}(f)|^2 df
    \label{eq:kinematic_energy}
\end{equation}
where $\mathbf{a}$ denotes discrete acceleration and $\mathcal{F}$ the spectral transform. 

Furthermore, we examine the frequency response of discrete acceleration to characterize temporal instability in acceleration space. Using central finite differences,
$\mathbf{a}_t=\mathbf{z}_{t+1}-2\mathbf{z}_t+\mathbf{z}_{t-1}$.
For a harmonic trajectory $\mathbf{z}_t=e^{i\omega t}$, $\omega\in[0,\pi]$, the resulting squared amplification factor is formally defined in Eqn.~\ref{eqn:amplify}.
\begin{equation}
A(\omega)=\frac{|\mathbf{a}_t|^2}{|\mathbf{z}_t|^2}
=4(1-\cos\omega)^2
\label{eqn:amplify}
\end{equation}
\noindent where $A(\omega)\!\to\!0$ as $\omega\!\to\!0$, while $A(\pi)=16$, indicating that discrete acceleration attenuates low-frequency motion and amplifies high-frequency variations. Compared with velocity, whose spectral gain is $2(1-\cos\omega)$, acceleration exhibits a sharper squared response of $4(1-\cos\omega)^2$. As a result, this makes the acceleration domain particularly suitable for temporal regularization, as it strongly penalizes high-frequency jitter and flickering while preserving inertial motion, since uniform linear motion has non-zero velocity but zero acceleration.
\subsection{Structured Autoregressive Noise Initialization}
\label{sub:structured_ar}
Motivated by the amplification analysis in Sec.~\ref{sub:acc_amplification}, we stabilize autoregressive generation at initialization by controlling adjacent-frame correlations in the noise trajectory. Since generated latents are recursively reused as causal context, these correlations can propagate and amplify temporal instability. We therefore construct the initial trajectory as a superposition of two independent, variance-preserving AR(1) processes with opposite temporal dynamics. This construction suppresses adjacent-frame correlations while preserving non-zero higher-lag correlations, as formally defined in Eqs.~\ref{eq:san} and~\ref{eq:san1}.
\begin{subequations}
\begin{align}
\mathbf{z}_t
&=
\cos\theta \, \mathbf{z}^{\mathrm{lf}}_t
+
\sin\theta \, \mathbf{z}^{\mathrm{hf}}_t, 
\label{eq:san}
\\
\mathbf{z}^{\mathrm{type}}_t
&=
\rho_{\mathrm{type}}
\mathbf{z}^{\mathrm{type}}_{t-1}
+
\sqrt{1-\rho_{\mathrm{type}}^2}
\,\boldsymbol{\epsilon}^{\mathrm{type}}_t 
\label{eq:san1}
\end{align}
\end{subequations}
where the innovation terms are mutually independent standard Gaussian noise. The temporal autocorrelation is $R(k)=\cos^2\theta,\rho_{\mathrm{lf}}^k+\sin^2\theta,\rho_{\mathrm{hf}}^k$. Setting $\theta=\frac{\pi}{4}$ and $\rho_{\mathrm{lf}}=-\rho_{\mathrm{hf}}=\rho$ yields $R(1)=0$ and $R(2)=\rho^2$, decorrelating adjacent noise frames while preserving non-zero higher-lag correlations. Importantly, $R(1)=0$ does not imply a high-frequency-dominated initialization. Instead, the symmetric low- and high-frequency components balance local temporal persistence and alternation, resulting in a kinematically neutral prior. By removing adjacent-frame temporal bias, SAN prevents the diffusion model from inheriting spurious local dependencies and provides a neutral starting point for subsequent guidance. Moreover, since the two independent autoregressive streams each have unit variance and are combined with variance-preserving weights, the resulting initialization remains standard Gaussian at each frame, preserving the marginal noise statistics expected by the pretrained diffusion model.

\subsection{Acceleration-Domain Spectral Guidance}
\label{sub:slepian_projection}
Although SAN (defined in Sec.~\ref{sub:structured_ar}) does not explicitly enforce motion smoothness at initialization, it provides a kinematically neutral baseline that mitigates short-term temporal bias. As denoising progresses, however, non-physical high-frequency perturbations can still accumulate during the autoregressive rollout. Building on the analysis in Sec.~\ref{sub:acc_amplification}, we therefore introduce a training-free guidance mechanism that regularizes the acceleration dynamics of the predicted clean latents. To obtain a reliable spectral estimate over short temporal windows while minimizing spectral leakage, we employ Discrete Prolate Spheroidal Sequences (DPSS), which maximize in-band energy concentration as defined in Eqn.~\ref{eq:slepian_concentration}.
\begin{equation}
\label{eq:slepian_concentration}
\max_{\mathbf{v}}
\frac{\int_{-W}^{W} |V(f)|^2  df}
{\int_{-1/2}^{1/2} |V(f)|^2 df}
\end{equation}
where $V(f)$ is the Fourier transform of $\mathbf{v}$ and $W$ the target half-bandwidth. DPSS improves spectral localization over short temporal windows, reducing leakage in acceleration-energy estimation. We therefore project $\mathbf{a}$ onto the DPSS basis $\{\mathbf{v}_k\}$ as $\beta_k=\mathbf{v}_k^\top\mathbf{a}$ and formally define the kinematic energy in Eqn.~\ref{eq:kin_slepian}.
\begin{equation}
\mathcal{E}_{\mathrm{kin}}(\hat{\mathbf{z}}_0)
=
\sum_{k=K_c}^{N-1}\|\beta_k\|^2,
\label{eq:kin_slepian}
\end{equation}
where $K_c$ denotes the cutoff index separating low- and high-frequency components. Rather than modifying model weights, we incorporate the kinematic prior through inference-time guidance directly on the latent factors defined in Eqn.~\ref{eq:guidance_z0}.
\begin{equation}
    \tilde{\mathbf{z}}_0 = \hat{\mathbf{z}}_0 - \eta \nabla_{\hat{\mathbf{z}}_0} \mathcal{E}_{\mathrm{kin}}(\hat{\mathbf{z}}_0),
    \label{eq:guidance_z0}
\end{equation}
where $\eta$ controls the guidance strength. By operating in acceleration-frequency space, SAGA suppresses high-frequency temporal instability while preserving low-frequency motion dynamics and the pretrained generative prior.

%% file: sec/Experiments.tex
\subsection{Experimental Setup} \label{sec:experimental_setup}
\input{sec/Experiment_setup}

\subsection{Quantitative Results}

\label{subsec:vbecnh2}

\noindent\textbf{Comparison with State-of-the-Art Methods.}
Table~\ref{tab:temporal_motion} compares SAGA with diffusion-based and autoregressive-diffusion video generation methods on VBench, 
showing consistent TQ and SC improvements across all three autoregressive backbones. In particular, the largest gain occurs on Self-Forcing, where TQ increases from 97.30 to 97.91 and IQ from 69.60 to 70.51, indicating improved temporal stability without sacrificing per-frame fidelity.
As a result, this suggests that the band-limited regularization in Sec.~\ref{sub:slepian_projection} improves temporal stability without indiscriminately smoothing the latent trajectory, thereby preserving low-frequency motion structure and per-frame visual fidelity.

\label{subsec:spectrum}
\begin{figure}[!t]
    \centering
    \includegraphics[width=\linewidth]{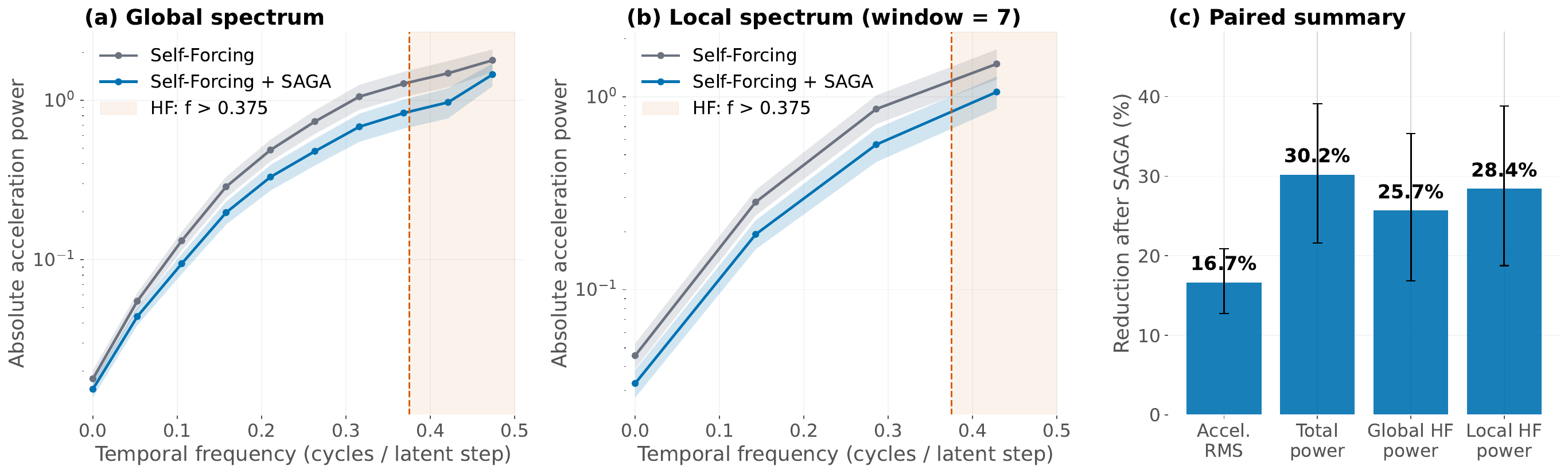}
    \caption{Global and local acceleration spectra over $100$ matched videos, together with paired reductions in acceleration RMS, total acceleration power, global high-frequency power, and local high-frequency power. Notably, the high-frequency band is formally defined as $f>0.375$, and error bars denote paired-bootstrap $95\%$ confidence intervals. \textbf{Best viewed in color and zoomed in.}}
    \label{fig:analysis}
\end{figure}

\noindent\textbf{Acceleration-Spectrum Analysis.}
To directly examine the mechanism underlying our proposed approach, we analyze the latent acceleration spectra of $100$ matched videos generated by Self-Forcing and Self-Forcing+SAGA. As shown in Fig.~\ref{fig:analysis}, SAGA reduces acceleration RMS by $16.7\%$ and total acceleration power by $30.2\%$, indicating fewer abrupt second-order temporal variations.

More importantly, the reduction is pronounced in the high-frequency range predicted by the amplification analysis in Sec.~\ref{sub:acc_amplification}. For $f>0.375$, SAGA reduces global and local high-frequency acceleration power by $25.7\%$ and $28.4\%$, respectively. This spectral response aligns with the guidance objective in Sec.~\ref{sub:slepian_projection} and supports our hypothesis that \emph{autoregressive rollout instability is associated with high-frequency acceleration components and that suppressing these components improves temporal stability}. The consistent global and local reductions further suggest that our proposed approach attenuates unstable dynamics throughout the rollout rather than only in isolated temporal segments. Since all comparisons use matched prompts and random seeds, the observed spectral changes can be attributed to the proposed guidance rather than prompt-level variation.

\begin{wrapfigure}[11]{r}{0.55\linewidth}
    \centering
    \includegraphics[
        width=\linewidth,
        trim=0 25 0 25,
        clip
    ]{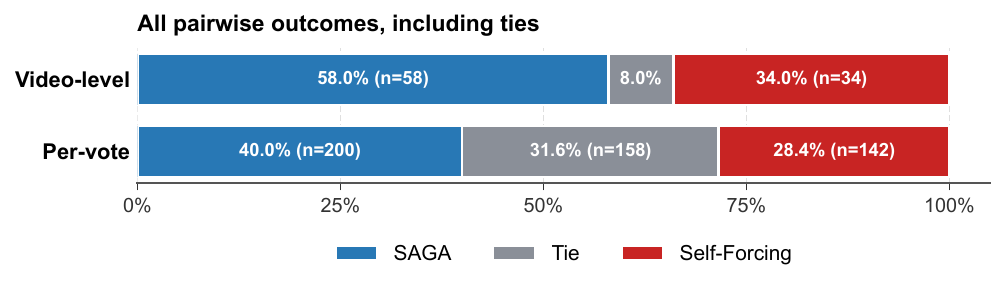}
    \caption{Human preference study with ties, reporting pairwise preferences between Self-Forcing and Self-Forcing+SAGA at the vote level and after video-level majority aggregation. \textbf{Best viewed in color and zoomed in.}}
    \label{fig:human_preference_with_ties}
\end{wrapfigure}
\noindent\textbf{Human Preference Study.}
\label{subsec:user} To assess whether the temporal improvements introduced by SAGA are perceptually noticeable, we conduct a blind pairwise user study on $100$ prompts sampled from MovieGenBench~\cite{polyak2024movie}. Ten participants compare randomized and anonymized video pairs generated by Self-Forcing~\cite{huang2025selfforcing} and Self-Forcing+SAGA. Each participant evaluates $50$ pairs, resulting in $500$ judgments in total, with every prompt assessed by at least three participants. For each pair, evaluators select the video with better temporal consistency and overall perceptual quality, or indicate a tie when neither video is clearly preferred.

As shown in Fig.~\ref{fig:human_preference_with_ties}, SAGA receives $200$ votes, compared with $142$ for Self-Forcing and $158$ ties. After prompt-level majority aggregation, SAGA wins on $58$ videos versus $34$, with only $8$ ties. Excluding ties, SAGA achieves $58.5\%$ of vote-level and $63.0\%$ of video-level preferences. The larger margin after aggregation suggests that the improvements are consistent across prompts rather than driven by isolated judgments. Although the number of ties indicates that many comparisons remain perceptually close, the overall preference for SAGA complements the automatic metrics and shows that improved rollout stability translates into perceptibly better temporal coherence and fewer temporal artifacts.


\begin{figure}[!t]
    \centering
    \includegraphics[width=\linewidth]{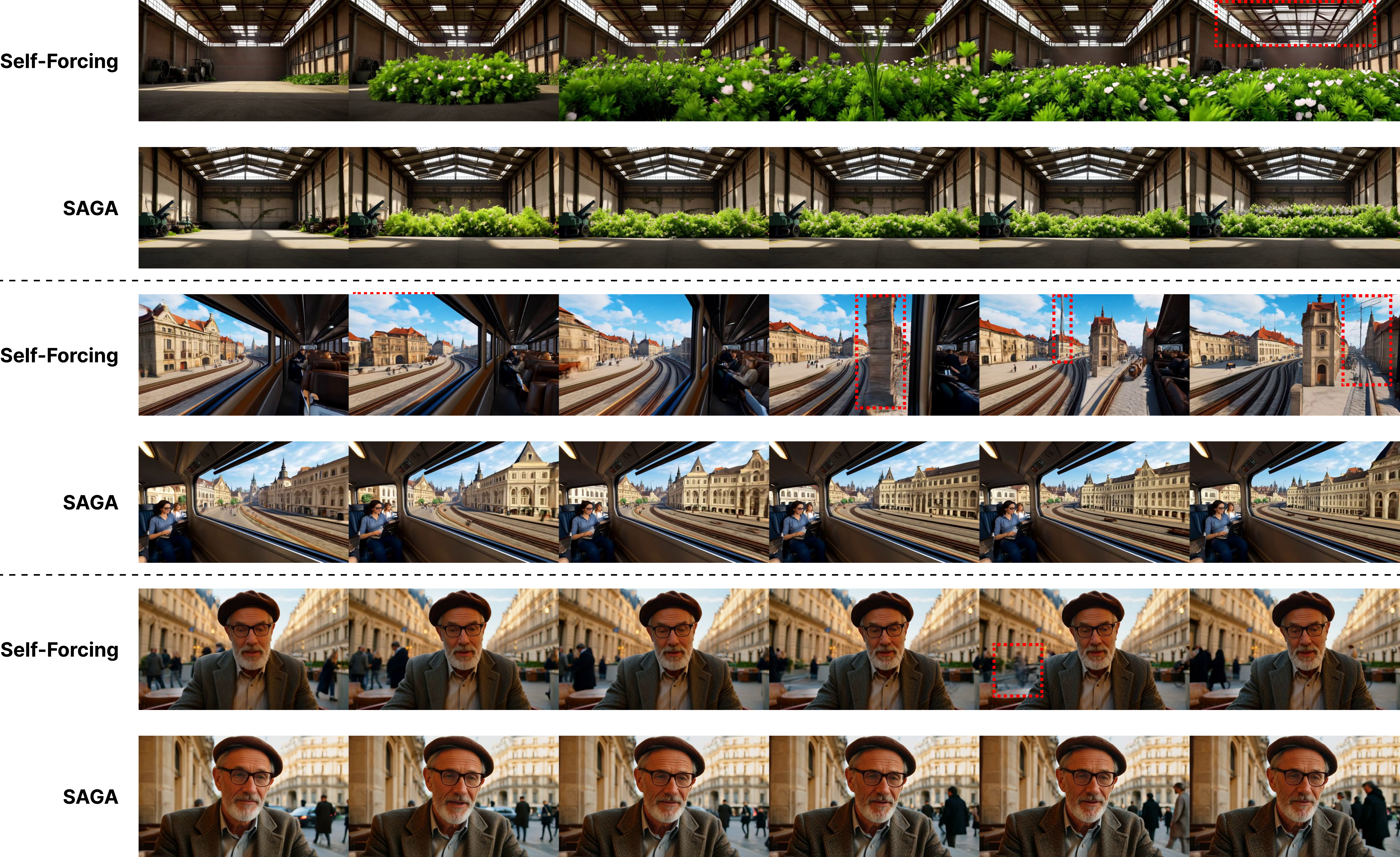}
    \caption{Qualitative comparison of Self-Forcing and Self-Forcing+SAGA using uniformly sampled frames. Red dashed boxes show temporal artifacts in the Self-Forcing baseline, including structural drift, background inconsistency, and local appearance flickering. \textbf{Best viewed in color and zoomed in.}}
    \label{fig:sample}
\end{figure}

\subsection{Qualitative Results}

Fig.~\ref{fig:sample} qualitatively compares Self-Forcing with and w/o SAGA on representative prompts. The baseline exhibits localized rollout instabilities, including structural drift, background variation, and transient distortions in moving or high-detail regions. In contrast, SAGA better preserves greenhouse geometry and plant structure, maintains more consistent scene boundaries and distant buildings under camera motion, and reduces appearance fluctuations around the human subject and surrounding crowd. These qualitative results suggest that SAGA suppresses local temporal perturbations before they accumulate into visible structural drift, consistent with the improvements in SC, BC, MS, and TQ reported in Table~\ref{tab:temporal_motion}.


%% file: sec/Experiment_setup.tex
\noindent\textbf{Datasets.}
We evaluate on 326 VBench prompts~\cite{huang2023vbench} across multiple autoregressive backbones, using identical prompts, seeds, and sampling schedules for paired comparisons. Ablations and human evaluation use a fixed 100-prompt subset of MovieGen Bench~\cite{polyak2024movie}, producing 100 videos per setting and 100 matched pairs.

\noindent\textbf{Evaluation Metrics.}
We evaluate four temporal dimensions from VBench: \textit{Subject Consistency} (SC) and \textit{Background Consistency} (BC), measured using cross-frame DINO~\cite{caron2021emerging} and CLIP~\cite{radford2021learning} similarity, respectively; \textit{Temporal Flickering} (TF), which measures short-term appearance variation; and \textit{Motion Smoothness} (MS), based on a pretrained frame-interpolation model~\cite{li2023amt}. We report their average as
$\mathrm{TQ}=(\mathrm{SC}+\mathrm{BC}+\mathrm{TF}+\mathrm{MS})/4$.
We additionally report \textit{Aesthetic Quality} (AQ) and \textit{Imaging Quality} (IQ)~\cite{ke2021musiq} to assess per-frame visual quality.

\noindent\textbf{Backbones.}
We evaluate SAGA on three 1.3B-parameter Wan2.1 autoregressive-diffusion backbones: CausVid~\cite{yin2025causvid}, Self-Forcing~\cite{huang2025selfforcing}, and Causal-Forcing~\cite{zhu2026causal}. All models use chunk-wise autoregressive rollout with three latent frames per block. SAGA is applied solely during inference, without updating backbone parameters.

\noindent\textbf{Inference Configuration.}
We generate $81$ RGB frames at $480\times832$ resolution (${\sim}5$s at $16$ FPS) using four denoising steps and a classifier-free guidance scale of $3.0$. For paired comparisons, each baseline and its SAGA-enhanced counterpart use identical prompts, seeds, and sampling schedules. In particular, we adopt a single global configuration across all prompts and backbones: guidance strength $\eta = 3$, AR coefficient $\rho = 0.5$, and DPSS time-bandwidth product $NW = 1.5$. Since approximately $2NW$ DPSS modes are well concentrated in $[-W,W]$~\cite{thomson1982spectrum}, we set $K_c = 3$ as the cutoff between low-frequency concentrated modes and high-frequency acceleration modes. All main experiments run on 1 $\times$ 80GB NVIDIA A100 GPU. Separately, on an RTX 5090, SAGA adds only 0.21\% latency overhead and 0.008 GiB peak allocated memory.

\begin{table}[!t]
\centering
\caption{
Comparison of diffusion-based and autoregressive-diffusion video generation models on the VBench benchmark~\cite{huang2023vbench}. Notably, SC, BC, TF, MS, TQ, AQ, and IQ correspond to Subject Consistency, Background Consistency, Temporal Flickering, Motion Smoothness, Temporal Quality, Aesthetic Quality, and Image Quality, respectively.
Best results are shown in \textbf{bold}, and second-best results are \underline{underlined}.}

\label{tab:temporal_motion}
\begin{tabular*}{\linewidth}{@{\extracolsep{\fill}}l|c|cccc|ccc}
\toprule
\textbf{Model} & \textbf{Para.} &\textbf{SC} & \textbf{BC} & \textbf{TF} & \textbf{MS} & \textbf{TQ}& \textbf{AQ} & \textbf{IQ}\\
\midrule

\rowcolor{gray!15} \multicolumn{9}{@{}l}{\textit{Diffusion Models}} \\
VideoCrafter-1.0~\cite{chen2023videocrafter1} &1.7B& 95.10 & \underline{98.04} & 98.93 & 95.67 & 96.94 & 62.67 & 65.46\\
CogVideoX~\cite{yang2024cogvideox} \textit{(ICLR 2025)}    &5.0B& \textbf{96.45} & 96.71 & \underline{98.97} & 97.20& 97.33 & 61.88 & 63.33 \\
Sora~\cite{liu2024sora}             &-& \underline{96.23} & 96.35 & 98.87 & \textbf{98.74} & 97.55 & 63.46 & 68.28\\
Wan2.1~\cite{wan2025}           &1.3B& 94.24&	\textbf{98.05}&	\textbf{99.25}&	\underline{98.15} &  97.64 & 62.43 & 66.51\\
\midrule

\rowcolor{gray!15} \multicolumn{9}{@{}l}{\textit{Autoregressive-Diffusion Hybrid Models }} \\
CausVid~\cite{yin2025causvid} \textit{(CVPR 2025)} &1.3B&95.91& 96.46& \underline{99.02}& 97.81 & 97.30&	64.18&	67.23\\
Self-Forcing~\cite{huang2025selfforcing} \textit{(NeurIPS 2025)}&1.3B& 95.60 & 96.20 & 99.00 & 98.38& 97.30 & 66.33 & 69.60  \\
Causal-Forcing~\cite{zhu2026causal} \textit{(ICML 2026)}&1.3B&95.43&	96.15&	97.97&	97.52&	96.77&	66.82&	\underline{70.11}\\
\midrule
\rowcolor{highlight}  \multicolumn{9}{@{}l}{\textit{Our approaches}} \\
\textbf{CausVid} + \textbf{SAGA}& 1.3B& \textbf{96.71}&  96.59&  98.96&  98.04&  97.58&  64.39&  67.46\\
 \textbf{Self-Forcing + SAGA} &1.3B& \underline{96.63}&	\textbf{96.95}&	\textbf{99.19}&	\textbf{98.85}& \textbf{97.91}&	\textbf{66.48}&	\textbf{70.51}\\
  \textbf{Causal-Forcing + SAGA}& 1.3B& 96.13&	96.57&	98.63&	97.97&	97.33&	66.66&	69.83\\
\bottomrule
\end{tabular*}%
\end{table}


%% file: sec/ablation.tex
\noindent\textbf{Effect of Key Components.} We isolate the contributions of structured autoregressive noise initialization (SAN; Sec.~\ref{sub:structured_ar}) and acceleration-domain spectral guidance (SG; Sec.~\ref{sub:slepian_projection}) on Self-Forcing. As shown in Table~\ref{tab:ablation_component}, SG alone improves Temporal Quality from $97.30$ to $97.58$, with consistent gains in Subject Consistency, Background Consistency, Temporal Flickering, and Motion Smoothness, supporting the analysis in Sec.~\ref{sub:acc_amplification} and Fig.~\ref{fig:analysis} that suppressing high-frequency acceleration dynamics improves rollout stability. However, its lower Aesthetic and Image Quality scores indicate a fidelity trade-off without a structured temporal prior. SAN alone increases Temporal Quality to $97.50$ and Image Quality to $70.03$, suggesting that structured inter-frame correlations provide a better-conditioned initialization while preserving visual fidelity. Combining SAN and SG yields the strongest performance, reaching $97.91$ Temporal Quality and $70.51$ Image Quality, consistent with Table~\ref{tab:temporal_motion} and the reduced temporal artifacts in Fig.~\ref{fig:sample}. These results demonstrate their complementary roles: SAN stabilizes the initial temporal trajectory, while SG suppresses residual high-frequency acceleration dynamics during denoising, jointly improving temporal coherence while mitigating the fidelity degradation observed with SG alone.

\begin{table}[!t]
\centering
\scriptsize
\caption {Ablation study of SAGA components, evaluating the individual contributions of structured autoregressive noise initialization (SAN) and spectral guidance (SG). Best results are shown in \textbf{bold}, and second-best results are \underline{underlined}.}
\label{tab:ablation_component}
\setlength{\tabcolsep}{6pt}
\begin{tabular}{cc|cccc|ccc}
\toprule
\textbf{SAN} & \textbf{SG} &\textbf{SC} & \textbf{BC} & \textbf{TF} & \textbf{MS} & \textbf{TQ}& \textbf{AQ} & \textbf{IQ}\\
\midrule

\xmark&\xmark& 95.60 & 96.20 & 99.00 & 98.38& 97.30 & 66.33 & 69.60\\
\xmark&\cmark& \underline{96.09}&	\underline{96.53}&	99.07&	\underline{98.62}&	\underline{97.58}&	66.15&	68.82\\
\cmark&\xmark& 95.87&	96.41&	\textbf{99.19}&	98.53&	97.50&	\textbf{66.59}&	\underline{70.03}\\
\midrule
\cmark&\cmark& \textbf{96.63}&	\textbf{96.95}&	\textbf{99.19}&	\textbf{98.85}& \textbf{97.91}&	\underline{66.48}&	\textbf{70.51}\\

\bottomrule
\end{tabular}%

\end{table}

\begin{table}[!t]
\centering

\caption{Ablation on the spectral decomposition basis used in acceleration-domain guidance. We compare the Fourier basis (FFT) with the Slepian basis constructed from DPSS. Best results are shown in \textbf{bold}, and second-best results are \underline{underlined}.}

\label{tab:fft_slepian}
\scriptsize
\begin{tabular*}{\linewidth}{@{\extracolsep{\fill}}l|c|cccc|ccc}
\toprule
\textbf{Model}&Basis&\textbf{SC} & \textbf{BC} & \textbf{TF} & \textbf{MS} & \textbf{TQ}& \textbf{AQ} & \textbf{IQ}\\
\midrule
Self-Forcing&--& 95.60 & 96.20 & 99.00 & 98.38& 97.30 & 66.33 & 69.60\\
\midrule
\textbf{Self-Forcing + SAGA} &FFT& \underline{96.60}&	\underline{96.91}&	\textbf{99.19}&	\textbf{98.85}&	\underline{97.89}&	\textbf{66.50}&	\underline{70.49}\\
\textbf{Self-Forcing + SAGA} &Slepian& \textbf{96.63}&	\textbf{96.95}&	\textbf{99.19}&	\textbf{98.85}& \textbf{97.91}&	\underline{66.48}&	\textbf{70.51}\\

\bottomrule
\end{tabular*}%
\end{table}
\begin{table}[!t]
\centering

\caption{
Evaluation of SAGA under two autoregressive rollouts: chunkwise and framewise generation. Performance is measured using VBench metrics~\cite{huang2023vbench}, including Subject Consistency (SC), Background Consistency (BC), Temporal Flickering (TF), Motion Smoothness (MS), Temporal Quality (TQ), Aesthetic Quality (AQ), and Image Quality (IQ).
Best results are shown in \textbf{bold}, and second-best results are \underline{underlined}.}
\scriptsize
\label{tab:rollout_granularity}
\begin{tabular*}{\linewidth}{@{\extracolsep{\fill}}l|cccc|ccc}
\toprule
\textbf{Model} &\textbf{SC} & \textbf{BC} & \textbf{TF} & \textbf{MS} & \textbf{TQ}& \textbf{AQ} & \textbf{IQ}\\
\midrule
\rowcolor{highlight}  \multicolumn{8}{@{}l}{\textit{Chunkwise AutoRegressive}} \\
Causal-Forcing&95.43&	96.15&	97.97&	97.52&	96.77&	\textbf{66.82}&	70.11\\

  \textbf{Causal-Forcing + SAGA}&  \textbf{96.13}&	\textbf{96.57}&	\textbf{98.63}&	\textbf{97.97}&	\textbf{97.33}&	\underline{66.66}&	\textbf{69.83}\\
\midrule
\rowcolor{highlight}  \multicolumn{8}{@{}l}{\textit{Framewise AutoRegressive}} \\
Causal-Forcing&\textbf{91.49}&	\textbf{93.12}&	\underline{94.64}&	\textbf{97.23}&	\textbf{94.12}&	\textbf{65.97}&	\underline{69.59}\\
  \textbf{Causal-Forcing + SAGA}&  \underline{91.23}&	\textbf{93.12}&	\textbf{94.66}&	\underline{97.20}&	\underline{94.05}&	\underline{65.85}	&\textbf{69.64}\\
\bottomrule
\end{tabular*}%
\end{table}

\noindent\textbf{Effect of Spectral Decomposition Basis.} We study the spectral basis used by the acceleration-domain guidance in Sec.~\ref{sub:slepian_projection}. Specifically, the main challenge is to separate low-frequency motion structure from the high-frequency acceleration components identified in Sec.~\ref{sub:acc_amplification} over short autoregressive windows. Although FFT provides a direct frequency decomposition, finite-window estimation suffers from spectral leakage and inadvertently penalizes valid low-frequency motion. Thus, we compare FFT- and Slepian-based guidance using the same backbones, prompts, random seeds, and sampling configurations.

As shown in Table~\ref{tab:fft_slepian}, both variants substantially outperform Self-Forcing, increasing Temporal Quality from $97.30$ to $97.89$ with FFT and $97.91$ with Slepian. This result complements the acceleration-spectrum analysis in Fig.~\ref{fig:analysis}, suggesting that the primary gain stems from suppressing high-frequency acceleration dynamics rather than from the choice of spectral basis. However, Slepian achieves the best Subject Consistency, Background Consistency, and Image Quality, while matching FFT in Temporal Flickering and Motion Smoothness.

\begin{wrapfigure}[17]{r}{0.45\textwidth}
  \centering
  \includegraphics[width=\linewidth]{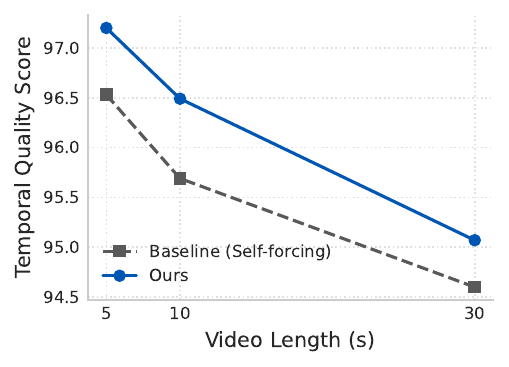}
  
  \caption{Long-horizon evaluation of Self-Forcing and Self-Forcing+SAGA (ours) at different video durations. \textbf{Best viewed in color and zoomed in.}}
  \label{fig:longvideo}
\end{wrapfigure}

Although the quantitative gap between FFT and Slepian is marginal, the consistent improvement of both variants over the baseline indicates that the primary contribution of SAGA lies in the acceleration-domain guidance framework rather than in the choice of spectral basis. Nevertheless, from a mathematical perspective, standard Fourier truncation is more susceptible to spectral leakage over short temporal windows, whereas Slepian sequences are designed to maximize spectral concentration for finite-length signals. We therefore do not treat DPSS as the main theoretical contribution, but retain Slepian as the default, principled implementation due to its optimal spectral concentration and negligible additional computational overhead.


\noindent\textbf{Effect of Autoregressive Rollout Granularity.} We study the effect of autoregressive rollout granularity on SAGA using Causal-Forcing as the backbone. Since structured initialization (introduced in Sec.~\ref{sub:structured_ar}) captures inter-frame dependencies and spectral guidance (introduced in Sec.~\ref{sub:slepian_projection}) estimates acceleration dynamics over short temporal windows, we compare chunk-wise and frame-wise rollouts to assess the role of temporal support in kinematic regularization. As shown in Table~\ref{tab:rollout_granularity}, SAGA consistently improves chunk-wise rollout, increasing Subject Consistency from $95.43$ to $96.13$, Background Consistency from $96.15$ to $96.57$, Temporal Flickering from $97.97$ to $98.63$, Motion Smoothness from $97.52$ to $97.97$, and Temporal Quality from $96.77$ to $97.33$. In contrast, under frame-wise rollout, SAGA yields only marginal changes, with Temporal Quality slightly decreasing from $94.12$ to $94.05$. This contrast suggests that SAGA benefits from multi-frame temporal support, which enables more reliable estimation and suppression of the high-frequency acceleration dynamics identified in Sec.~\ref{sub:acc_amplification}. Consistent with the spectral analysis in Fig.~\ref{fig:analysis}, these results further indicate that acceleration-domain regularization is most effective under chunk-wise rollout, where multiple frames within each autoregressive block provide the local temporal context required for reliable kinematic guidance.

\noindent\textbf{Long-Horizon Video Generation.} AR video generation becomes increasingly challenging over long rollout horizons, as local temporal perturbations can propagate through recursively generated causal context. To assess SAGA under extended rollout, we compare Self-Forcing~\cite{huang2025selfforcing} and Self-Forcing+SAGA at durations of $5$, $10$, and $30$ seconds using VBench Temporal Quality~\cite{huang2023vbench}.
As shown in Fig.~\ref{fig:longvideo}, temporal quality decreases with increasing rollout length for both methods, but SAGA consistently outperforms Self-Forcing from 5 to 30 seconds. The persistent gain suggests that SAGA mitigates, though does not eliminate, temporal error accumulation over extended autoregressive rollout.

%% file: sec/conclusion.tex
In this work, we have presented SAGA, a training-free framework for stabilizing autoregressive video diffusion through acceleration-domain spectral guidance and structured stochastic initialization. By targeting high-frequency perturbations amplified in latent acceleration space, SAGA improves rollout stability while preserving low-frequency motion structure and the pretrained generative prior. Experiments across multiple backbones demonstrate consistent temporal gains without retraining or architectural changes, while ablations, spectral analysis, and human evaluation validate the complementary design and perceptual benefits. Our findings highlight latent acceleration dynamics as an effective target for stabilizing autoregressive video generation. Future work will explore adaptive spectral regularization and its extension to longer-horizon generation.

\section*{Acknowledgement}
This research is funded by the Vietnam National Foundation for Science and Technology Development (NAFOSTED) under grant number 102.01-2025.66. Trong-Thuan Nguyen was funded by the PhD Scholarship Programme of Vingroup Innovation Foundation (VINIF), VinUniversity, code VINIF.2025.TS63.